\documentclass[nohyperref]{article}

\usepackage{microtype}
\usepackage{graphicx}
\usepackage{subfigure}
\usepackage{booktabs} %

\usepackage{hyperref}

\usepackage[accepted]{icml2023}

\usepackage{amsmath}
\usepackage{amssymb}
\usepackage{mathtools}
\usepackage{amsthm}

\usepackage[capitalize,noabbrev]{cleveref}

\theoremstyle{plain}

\theoremstyle{definition}

\theoremstyle{remark}

\usepackage[textsize=tiny]{todonotes}

\newcommand{\Tbl}[1]  {Table \ref{tbl:#1}}

\newcommand{\methodName}{CHOIR\xspace}
\newcommand{\ie}{\textit{i.e.}\xspace}
\newcommand{\eg}{\textit{e.g.}\xspace}

\usepackage{xr}
\makeatletter
\newcommand*{\addFileDependency}[1]{%
  \typeout{(#1)}
  \@addtofilelist{#1}
  \IfFileExists{#1}{}{\typeout{No file #1.}}
}
\makeatother

\usepackage{graphicx}
\usepackage{amsmath}
\usepackage{amssymb}
\usepackage{booktabs}
\usepackage{dsfont}
\usepackage{mathtools}
\usepackage{multirow}
\usepackage{graphics}
\usepackage{wrapfig}
\usepackage{bbm}
\usepackage{mwe}
\usepackage{xspace}

\icmltitlerunning{Stable and Consistent Prediction of 3D Characteristic Orientation via Invariant Residual Learning}

\begin{document}

\twocolumn[
\icmltitle{Stable and Consistent Prediction of 3D Characteristic Orientation\\ via Invariant Residual Learning}  

\icmlsetsymbol{equal}{*}

\begin{icmlauthorlist}
\icmlauthor{Seungwook Kim}{equal,gsai}
\icmlauthor{Chunghyun Park}{equal,gsai}
\icmlauthor{Yoonwoo Jeong}{gsai}
\icmlauthor{Jaesik Park}{gsai,cse}
\icmlauthor{Minsu Cho}{gsai,cse}

\end{icmlauthorlist}

\icmlaffiliation{gsai}{Graduate School of Artificial Intelligence, POSTECH, Pohang, South Korea}
\icmlaffiliation{cse}{Department of Computer Science and Engineering, POSTECH, Pohang, South Korea}

\icmlcorrespondingauthor{Minsu Cho}{mscho@postech.ac.kr}

\icmlkeywords{Characteristic orientation, Canonical orientation, Self-supervised learning, SO(3)-Equivariance, Point cloud}

\vskip 0.3in
]

\printAffiliationsAndNotice{\icmlEqualContribution} %

\begin{abstract}

Learning to predict reliable characteristic orientations of 3D point clouds is an important yet challenging problem, as different point clouds of the same class may have largely varying appearances.
In this work, we introduce a novel method to decouple the shape geometry and semantics of the input point cloud to achieve both stability and consistency.
The proposed method integrates shape-geometry-based SO(3)-equivariant learning and shape-semantics-based SO(3)-invariant residual learning, where a final characteristic orientation is obtained by calibrating an SO(3)-equivariant orientation hypothesis using an SO(3)-invariant residual rotation.
In experiments, the proposed method not only demonstrates superior stability and consistency but also exhibits state-of-the-art performances when applied to point cloud part segmentation, given randomly rotated inputs. 

\end{abstract}

\section{Introduction}
\label{sec:intro}

With recent advances in deep learning, there have been successful attempts to reason about 3D geometric data such as point clouds or 
meshes~\cite{charles17pointnet,dgcnn,pointconv}.
These attempts have eschewed the need to rely on handcrafted features~\cite{weinmann15handcrafted} to perform various tasks, including 3D object recognition, 3D semantic segmentation, and point cloud registration.
However, a consistent inference with respect to variations of input point clouds remains challenging, especially when an arbitrary rotation is involved in the point clouds to handle.

While there exist methods to yield rotation-robust representations by employing rotation-equivariant~\cite{shen20203d,deng2021vector} or rotation-invariant~\cite{srinet2019, Yu2020DeepPA, li2021rotation} networks, the yielded representations suffer from loss of expressivity in return for robustness to rotation. 
A more straightforward and intuitive alternative is to \textit{canonicalize} the point clouds by determining their \textit{characteristic} orientation~\cite{fang2020rotpredictor,katzir2022shape}; the canonicalized point cloud, which is obtained by canceling out the characteristic orientation, preserves its shape information and is readily used for downstream tasks without loss of expressivity.
Although it is non-trivial to define the canonical orientation of a point cloud, there exist its desiderata~\cite{katzir2022shape}: \textit{stability} - the canonical orientation is invariant to any rotation transformations of the input point cloud - and \textit{consistentency} - the canonical orientation is invariant to intra-class variations \ie, point clouds of the same class should have the same canonical pose.
Recent methods to canonicalize the point clouds commonly exploit these desiderata to train their network in a self-supervised manner~\cite{sun2021canonical, sajnani2022condor,katzir2022shape}.

Existing methods to predict the characteristic orientation of point clouds, however, suffer from satisfying both of the criteria, stability and consistency, simultaneously.
Earlier methods either rely on heavy rotation augmentation~\cite{sun2021canonical} to model equivariance, or rely on networks that are not purely equivariant~\cite{spezialetti2020learning} to predict characteristic orientations, which therefore yield suboptimal stability.
More recent methods propose to disentangle the shape and pose of point clouds by yielding rotation-invariant shape representation and rotation-equivariant pose information~\cite{sajnani2022condor, katzir2022shape}.
These methods are tightly coupled with a class-sensitive point cloud reconstruction pipeline~\cite{groueix2018papier}, consequently making them unsuitable for multi-class training.
While they may be stable, such approaches often fail to obtain consistent orientation predictions.

In this work, we introduce an effective method to learn the characteristic orientation of point clouds achieving both stability and consistency. The idea is to decouple the shape information of a point cloud into two factors, pure geometry and semantics, and coordinate them by means of SO(3)-equivariant learning and SO(3)-invariant residual learning.
Specifically, we predict a SO(3)-equivariant characteristic orientation hypothesis and calibrate it using SO(3)-invariant residual rotation to yield a final prediction for stable and consistent canonicalization. We name the proposed CHaracteristic Orientation predictor with Invariant Residual learning as \methodName.
As it dispenses with the widely used point cloud reconstruction in learning~\cite{sajnani2022condor,katzir2022shape}, \methodName facilitates multi-class training, being more practical for downstream point cloud analysis tasks.

The contribution of our work is fourfold. 
\textbf{(1)} We introduce \methodName, a strong characteristic orientation prediction method that decouples shape information of point clouds to geometry and semantics to achieve both stability and consistency.
\textbf{(2)} We propose a novel residual predictor that learns to output a SO(3)-invariant residual rotation to calibrate our initial orientation hypotheses for better consistency.
\textbf{(3)} In experiments, we not only demonstrate state-of-the-art stability and consistency, but also show superior performance when applied to the task of point cloud part segmentation.
\textbf{(4)} We facilitate multi-class training by removing point cloud reconstruction, enhancing applicablity to downstream tasks,
\textbf{(5)} We analyze the influence of consistency and stability on downstream tasks, which indicates that the balance between stability and consistency is the key to canonicalizing point clouds for downstream tasks.

\section{Related Work}
\label{sec:related_work}

\smallbreak
\noindent
\textbf{Learning based point cloud analysis.}
Over the recent years, various methods have been proposed to handle point clouds, which pose difficulties due to their irregularity and lack of order.
PointNet~\cite{charles17pointnet} and its follow-up methods~\cite{qi17pointnetpp, pointconv, thomas2019kpconv, dgcnn} propose to input point clouds directly to networks to achieve permutation-invariant representations of the point cloud.
For example, DGCNN~\cite{dgcnn} uses graph convolution using pointwise k-nearest-neighbors, while KPConv~\cite{thomas2019kpconv} defines the convolution area using a set of kernel points.
More recently, Point Transformer~\cite{zhao2021point} proposed to leverage the transformer architecture to process point clouds, while PointMixer~\cite{choe2022pointmixer} facilitates information sharing between points by replacing token-mixing MLPs with Softmax function.
While these methods have shown impressive performances on point cloud analysis tasks including classification and segmentation, these methods 
rely on extensive train-time or test-time rotation augmentations to handle random rotations encountered during inference.

\smallbreak
\noindent
\textbf{Rotation-robust point cloud analysis.}
To minimize the need for intensive rotation augmentations, one thread of work aims to design rotation-equivariant modules, such that the output representations will be equivariant with respect to the rotation of the input point cloud~\cite{esteves2018learning, rao2019spherical, shen20203d, fang2020rotpredictor, thomas2018tensor, chen2021equivariant}.
To introduce a few recent successes, YOHO~\cite{wang2022you} exploits the symmetry of icosahedral group to extract group-equivariant representations.
\citet{luo2022equivariant} propose the idea of equivariant message passing by learning point-wise orientations.
Vector Neuron Networks~\cite{deng2021vector} lift the latent representations from scalar vectors to matrix vectors to implement a fully equivariant network.

Another thread of work explores achieving strict rotational invariance \ie, yielding the same output representation regardless of the rotation of the input point cloud.
Many existing work focus on handcrafting rotation-invariant features based on intrinsic geometries~\cite{srinet2019,li2021rotation, xiao2021triangle}.
While the above methods suffer from information loss, some approaches propose to leverage PCA-based canonical poses to achieve rotational invariance~\cite{kim20rigcn, Yu2020DeepPA, li2021closer}.
However, PCA-based canonical poses suffer from sign and order ambiguities~\cite{li2021closer}, and considering all the possible candidates for the canonical pose leads to high computational overhead. %

In our work, we propose to leverage SO(3)-equivariant networks for learning the shape geometry, and SO(3)-invariant networks for learning the shape semantics, to predict the characteristic orientation of point clouds.
Canonicalizing the input point cloud using our predicted characteristic orientation enables us to preserve the shape information without having to consider various ambiguities.

\smallbreak
\noindent
\textbf{Self-supervised category-level 6D pose estimation.}
The task of category-level pose estimation aims to find the 6D poses of unseen instances from known categories, without access to object CAD models.
A similar approach called shape co-alignment~\cite{averkiou2016autocorrelation, mehr2018manifold} proposes to obtain consistent category-level reference frames, which require clean mesh inputs or rely heavily on the symmetry of point clouds.
QEC~\cite{zhao2020quaternion} proposes a novel dynamic routing procedure on quarternions to establish end-to-end transformation equivariance, but QEC does not evaluate the predicted category-level pose, and assumes complete shapes.
EquiPose~\cite{li2021leveraging} proposes to handle both complete and partial shapes, and introduces a self-supervised learning framework to estimate several category-level object pose candidates from single 3D point clouds by leveraging EPNs~\cite{chen2021equivariant}.

While characteristic orientation prediction and 6D pose estimation tasks are pertinent, 6D pose estimation, unlike characteristic orientation prediction, has ground-truth category-level pose annotations that uniquely define a category-level pose reference frame.

\smallbreak
\noindent
\textbf{Estimation of characteristic orientation.}
Previously, some work proposed to learn the canonical frame of objects using strong or weak supervision~\cite{rempe2020caspr, novotny2019c3dpo, gu2020weakly}.
More recent methods propose to determine the characteristic orientation of point clouds in a self-supervised manner.
{Compass~\cite{spezialetti2020learning}} projects point clouds to spherical signals to be processed by SO(3) group-equivariant spherical CNNs, and leverages the arg-max of SO(3)-equivariant representations to predict the characteristic orientation of point clouds.
In {Canonical Capsules~\cite{sun2021canonical}}, a network is trained to decompose point clouds into parts, on which invariance/equivariance is enforced through a Siamese training setup. 
The point clouds are then canonicalized to a learned frame of reference by reconstruction.
{ConDor~\cite{sajnani2022condor}} leverages Tensor Field Networks (TFNs)~\cite{thomas2018tensor}, a class of permutation- and rotation equivariant, and translation-invariant 3D networks to learn to output an equivariant canonical pose.
{VN-SPD~\cite{katzir2022shape}} learns to disentangle shape and pose information from input point clouds by leveraging SO(3)-equivariant and SO(3)-invariant representations produced from Vector Neuron Networks~\cite{deng2021vector}.
Note that Canonical Capsules, ConDor and VN-SPD rely on point cloud reconstruction pipelines to obtain the canonicalized point clouds, making their model suitable only for a single class of point clouds.

In this work, we eschew the point cloud reconstruction pipeline and rather focus on learning the residual rotation using SO(3)-invariant layers to calibrate the initial characteristic orientation hypothesis obtained from our SO(3)-equivariant networks (that directly takes point clouds as the input), enforcing better consistency.

\section{Problem Statement}
\label{sec:problem_setup}

Given an input point cloud $\mathbf{P} \in \mathbbm{R}^{N\times 3}$, which is centered at the origin, we aim to predict its {\em characteristic} orientation $R' \in \mathrm{SO}(3)$ such that point clouds of the same class with {\em arbitrary rotation} and {\em intra-class variation} are all aligned to the same coordinate frame, called the {\em canonical} orientation, by canceling out the characteristic orientation via rotation $R'^\top$. 
The resultant canonicalized point cloud $\mathbf{P}R'^\top$ provides a rotation-invariant representation, which is crucial for robust 3D perception in practical scenarios. 
Since the main challenge of the problem lies in achieving the robustness to both (1) arbitrary rotation and (2) intra-class variation,   
the characteristic orientation prediction is commonly evaluated with two metrics: {\em stability} and {\em consistency}~\cite{sun2021canonical, katzir2022shape}.

The stability metric quantifies how similar canonical orientations are recovered from point clouds of the same instance with different orientations, \ie, the robustness to arbitrary rotation.  
Specifically, from a point cloud $\mathbf{P}$, randomly rotated ones $\{\mathbf{P} R_i\}$ are generated and then their characteristic orientations $\{ R_i'\}$ are predicted. As canonicalized point clouds correspond to $\{\mathbf{P} R_i R_i'^\top\}$, the stability metric is measured by the standard deviation of the net rotations $\{R_i R_i'^\top\}$:
\begin{equation}
\label{eq:stability}
d_\textrm{stability}(\{R_i R_i'^\top\}_{i=1}^K) = \sqrt{\frac{1}{K} \sum_{i=1}^{K} \angle (R_i R_i'^\top, \bar{R}) ^ 2},
\end{equation}
where $\bar{R}$ is the chordal $L_2$-mean of $\{R_i R_i'^\top\}_{i=1}^K$ and $\angle(\cdot, \cdot)$ denotes the angle difference between two rotation matrices.

On the other hand, the consistency metric quantifies how similar canonical orientations are recovered from different instances of the same category, \ie, the robustness to intra-class variation.
Specifically, given $N$ different point cloud instances from the same class, we measure the standard deviation of their predicted  orientations as follows:
\begin{equation}
\label{eq:consistency}
d_\textrm{consistency}(\{R'_j\}_{j=1}^N) = \sqrt{\frac{1}{N} \sum_{j=1}^{N} \angle (R'_j, \bar{R}) ^ 2},
\end{equation}
where $\bar{R}$ denotes the chordal $L_2$-mean of the predicted characteristic orientations $\{R'_j\}_{j=1}^N$.

\begin{figure*}[!t]
    \centering
    \resizebox{\textwidth}{!}{
    \includegraphics{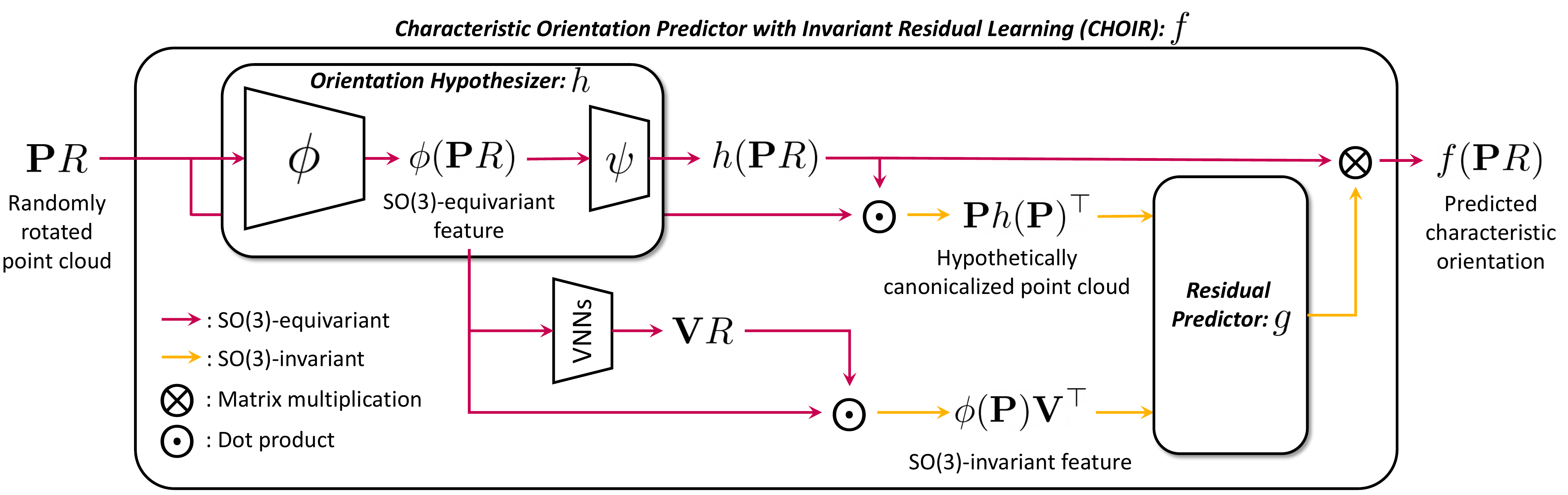}
    }
    \vspace{-6mm}
    \caption{\textbf{Overview of \methodName.} 
    \methodName ($f$) predicts the characteristic orientation using an SO(3)-equivariant orientation hypothesizer ($h$) and an SO(3)-invariant residual predictor ($g$). The orientation hypothesizer consists of two modules: SO(3)-equivariant encoder ($\phi$) and SO(3)-equivariant rotation predictor ($\psi$).
    The hypothesizer first predicts an equivariant characteristic orientation hypothesis, which is calibrated using the rotation-invariant residual rotation predicted by the residual predictor. 
    With this novel rotation-invariant residual learning, \methodName predict the characteristic orientation of a rotated point cloud that is both stable under arbitrary rotations and consistent under intra-class shape deformations.
    }
    \label{fig:overview}
\end{figure*}
\section{Our Approach}
\label{sec:method}

We introduce a characteristic orientation predictor, \methodName, that effectively minimizes both stability and consistency. %

The main idea of our approach is to identify two distinctive factors in determining a characteristic orientation and decouple them in learning and prediction. 
The first and primary factor is the pure shape geometry: a characteristic orientation may be assigned purely based on its shape geometry such that the rotated shape of the object has a consistent characteristic orientation corresponding to the rotation intact. 
The second factor is the shape semantics: a characteristic orientation may be affected by recognizing a specific semantic property, \eg, the signature of a specific object instance or class.  
Note that the shape semantics is of course related to the shape geometry but is distinct in the sense that it involves additional semantic information in learning. %
As can easily be noticed, these two factors are closely connected to the two criteria of characteristic orientation, stability and consistency, respectively, which may cause conflict with each other in learning.  
To achieve both stability and consistency, we decouple the two factors, shape geometry and shape semantics, and integrate them in a compatible manner by combining shape-geometry-based SO(3)-equivariant learning and shape-semantics-based SO(3)-invariant residual learning. Figure~\ref{fig:overview} illustrates the overview of our approach.

To guarantee the stability of predictions, we leverage SO(3)-equivariance in our network to predict a shape-geometry-based characteristic orientation hypothesis, such that it remains \emph{equivariant} to the rotation of the input~(Sec.~\ref{subsec:hypothesizer}). 
To enforce consistency of predictions, we introduce a SO(3)-invariant residual rotation predictor~(Sec.~\ref{subsec:residual-predictor}), which refines the characteristic orientation hypothesis by predicting a shape-semantics-based residual rotation from SO(3)-invariant features. 
Note that our residual strategy for consistency allows the final prediction to preserve the stability since SO(3)-invariant rotation does not hurt SO(3)-equivariance of the orientation hypothesis.
In training \methodName, the adoption of SO(3)-equivariant network for the characteristic orientation hypothesis intrinsically reflects the shape geometry, enforcing the stability by design, and thus we focus on learning the consistency based on shape semantics. To this end, we use a Siamese training setup on aligned point cloud pairs of different point clouds within the same class, using self-supervised losses to facilitate consistency between the predicted characteristic orientations of the point cloud pair (Sec.~\ref{subsec:self-sup}).

\subsection{Preliminary: Vector Neuron Networks}
In designing \methodName, one of our aims is to improve the stability of its characteristic orientation predictions.
To this end, we employ SO(3)-equivariant layers into our network architectures, inspired by Vector Neuron Networks (VNNs) ~\cite{deng2021vector} for their simplicity and generalizability.
In VNNs, the sequence of scalar values, which represents a single neuron, is lifted to a sequence of 3D vectors, \ie, a \emph{vector-list} feature $\mathbf{V} \in \mathbbm{R}^{C \times 3}$.
The layers of VNNs map from and to batches of such vector-list features such that $f(\mathbf{V}R) = f(\mathbf{V})R$   satisfying the equivariance to the rotation $R\in\mathrm{SO}(3)$.
SO(3)-invariant features can also be obtained using the product of an equivariant vector-list feature $\mathbf{V}R \in \mathbbm{R}^{C \times 3}$ and the transpose of another equivariant vector-list feature $\mathbf{U}R \in \mathbbm{R}^{C' \times 3}$: $(\mathbf{V}R)(\mathbf{U}R)^\top = \mathbf{V}RR^\top\mathbf{U}^\top = \mathbf{V}\mathbf{U}^\top$.
We refer the readers to the original  paper~\cite{deng2021vector} for more details.

\subsection{SO(3)-equivariant Orientation Hypothesizer}
\label{subsec:hypothesizer}

The characteristic orientation hypothesizer builds on the top of SO(3)-equivariant VNNs~\cite{deng2021vector} so that it initially predicts a SO(3)-equivariant orientation in our approach. Specifically, the characteristic orientation hypothesizer ($h~:=~\psi~\circ~\phi$) consists of two SO(3)-equivariant modules, an encoder $\phi: \mathbb{R}^{N\times3}\mapsto\mathbb{R}^{N\times C\times 3}$ and a rotation predictor $\psi~:~\mathbb{R}^{N\times C\times 3}\mapsto \mathrm{SO}(3)$.
The encoder processes a rotated point cloud $\mathbf{P}R \in \mathbbm{R}^{N\times3}$ using VNNs and extracts a SO(3)-equivariant feature $\phi(\mathbf{P}R)$.
Then, the SO(3)-equivariant rotation predictor, $\psi$, predicts an orientation hypothesis by reducing the channel of $\phi(\mathbf{P}R)$ from $C$ to $2$ to estimate two basis vectors of the orientation and orthonormalizing the vectors by Gram-Schmidt process. Therefore, our hypothesizer maintains SO(3)-equivariance as:
\begin{equation}
    \begin{split}
            h(\mathbf{P}R) &= \psi(\phi(\mathbf{P}R)) = \psi(\phi(\mathbf{P}))R =h(\mathbf{P})R.
    \end{split}
\end{equation}
Attributing to these properties of our hypothesizer, our characteristic orientation hypotheses are stable.

\subsection{SO(3)-invariant Residual Predictor}
\label{subsec:residual-predictor}
Due to the presence of intra-class shape variations within a class, our hypothesizer network alone is insufficient to obtain consistent class-specific characteristic orientations in spite of our usage of cross-shape point cloud pairs.
To this end, we introduce a residual orientation predictor that aims to calibrate the class-specific characteristic orientation hypothesis $h(\mathbf{P}R)$ such that it will be consistent within semantic classes.
Formally put, we assume there exists a true class-specific canonical orientation, and there is a residual between the true class-specific canonical orientation and the coordinate frame aligned by the predicted SE(3)-equivariant characteristic orientation hypothesis due to its SO(3)-invariant properties such as shape or partiality. Therefore, we introduce a SO(3)-invariant residual predictor ($g: \mathbbm{R}^{N\times 3} \times \mathbbm{R}^{N\times C\times 3}  \rightarrow \mathrm{SO}(3)$) to predict the residual and to align point clouds from the same class into the true class-specific canonical orientation.

We employ Point Transformer~\cite{zhao2021point}, which takes 3D points and their corresponding features as inputs, as our SO(3)-invariant residual predictor denoted by $g(\cdot, \cdot)$; and because we want the predicted residual to be invariant to rotations, we provide rotation-invariant features and a \textit{hypothetically} canonicalized point cloud as inputs.
We first generate the rotation-invariant feature using the intermediate SO(3)-equivariant feature $\phi(\mathbf{P}R)\in\mathbb{R}^{N\times C\times 3}$ from the hypothesizer network $h$. 
Specifically, we pass $\phi(\mathbf{P}R)$ through more SO(3)-equivariant layers to yield another SO(3)-equivariant representation $\mathbf{V}R\in \mathbbm{R}^{N\times 3\times3}$.
We then compute their inner product along the batch dimension $N$ to obtain rotation-invariant features of the input point cloud:
\begin{equation}
    \phi(\mathbf{P}R)(\mathbf{V}R)^\top= \phi(\mathbf{P})R R^\top\mathbf{V}^\top = \phi(\mathbf{P})\mathbf{V}^\top.
\end{equation}
To obtain the \textit{hypothetically} canonicalized point cloud, we apply our  orientation hypothesis onto the input: %
\begin{equation}
    \mathbf{P}Rh(\mathbf{P}R)^\top=\mathbf{P}R R^\top h(\mathbf{P})^\top = \mathbf{P}h(\mathbf{P})^\top.
\end{equation}
Since all inputs of the residual predictor $g(\cdot, \cdot)$ are SO(3)-invariant, we can predict a SO(3)-invariant residual orientation nonetheless even though we use Point Transformer~\cite{zhao2021point}, which is not SO(3)-invariant, as our residual predictor. 

With the SO(3)-equivariant orientation hypothesizer $h$ and the SO(3)-invariant residual predictor $g$, we finally construct a SO(3)-equivariant network $f:\mathbb{R}^{N\times 3} \mapsto \mathrm{SO}(3)$ to predict the characteristic orientation of the point cloud $\mathbf{P}R$ as:
\begin{equation}
\begin{split}
f(\mathbf{P}R) &= g(\mathbf{P}Rh(\mathbf{P}R)^\top, \phi(\mathbf{P}R)(\mathbf{V}R)^\top )h(\mathbf{P}R) \\
   &= g(\mathbf{P}h(\mathbf{P})^\top , \phi(\mathbf{P})\mathbf{V}^\top )h(\mathbf{P}R) \\
   &= g(\mathbf{P}h(\mathbf{P})^\top , \phi(\mathbf{P})\mathbf{V}^\top )h(\mathbf{P})R \\
   &= f(\mathbf{P})R.
\end{split}
\end{equation}
Figure~\ref{fig:overview} illustrates the overview of $f$.

\subsection{Self-Supervised Objective}
\label{subsec:self-sup}
Given two rotated point clouds of the same class, $\mathbf{P}_1 R_1$ and $ \mathbf{P}_2 R_2$, their respective outputs from \methodName \textit{i.e.,} characteristic orientations would be $R'_1=f(\mathbf{P_1}R_1)$ and $R'_2=f(\mathbf{P_2}R_2)$, respectively.
To achieve the consistency of the predicted characteristic orientations, we formulate our self-supervised objective ($\mathcal{L}$) as follows:
\begin{equation}
\begin{split}
\mathcal{L} &= \lVert {R'_1}^\top R'_2 - R_1^\top R_2 \rVert_F ^2\\
  &= \lVert {f(\mathbf{P}_1 R_1)}^\top f(\mathbf{P}_2 R_2) - R_1^\top R_2 \rVert_F ^2\\
  &= \lVert R_1^\top {f(\mathbf{P}_1)}^\top f(\mathbf{P}_2 )R_2 - R_1^\top R_2 \rVert_F ^2\\
  &= \lVert R_1^\top ({f(\mathbf{P}_1)}^\top f(\mathbf{P}_2 ) - I) R_2 \rVert_F ^2\\
  &= \lVert {f(\mathbf{P}_1)}^\top f(\mathbf{P}_2 ) - I \rVert_F ^2\\
  &= \lVert {f(\mathbf{P}_2)} - f(\mathbf{P}_1 ) \rVert_F ^2,
\end{split}
\end{equation}
where $I\in \mathbb{R}^{3\times 3}$ is the identity matrix.
In essence, minimizing $\mathcal{L}$ enforces $f$ to predict \emph{consistent} characteristic orientations from two different point clouds $\mathbf{P}_1$ and $\mathbf{P}_2$. It is worth noting that we do not use the pre-aligned coordinate frame of a dataset, which is aligned by \textit{humans}, as a ground truth canonical orientation to \textit{fully} supervise $f$ to align input point clouds to the ground truth.
Instead, we construct a pair of point clouds using either augmentation (\eg, $k$NN patch removal, resampling) or cross-instance sampling within the same class, and \textit{self}-supervise the class-specific canonical orientation, which may differ from the pre-aligned coordinate frame.

\section{Experiments}
\label{sec:experiments}
We perform qualitative and quantitative experiments to demonstrate the efficacy of our proposed method on the task of predicting the characteristic orientations of point clouds.
Also, we analyze the significance of stability and consistency of characteristic orientation estimation on a downstream point cloud analysis task, the ShapeNet~\cite{shapenet} part segmentation task.

\begin{figure*}[!t]
    \centering
    \resizebox{\textwidth}{!}{
    \includegraphics{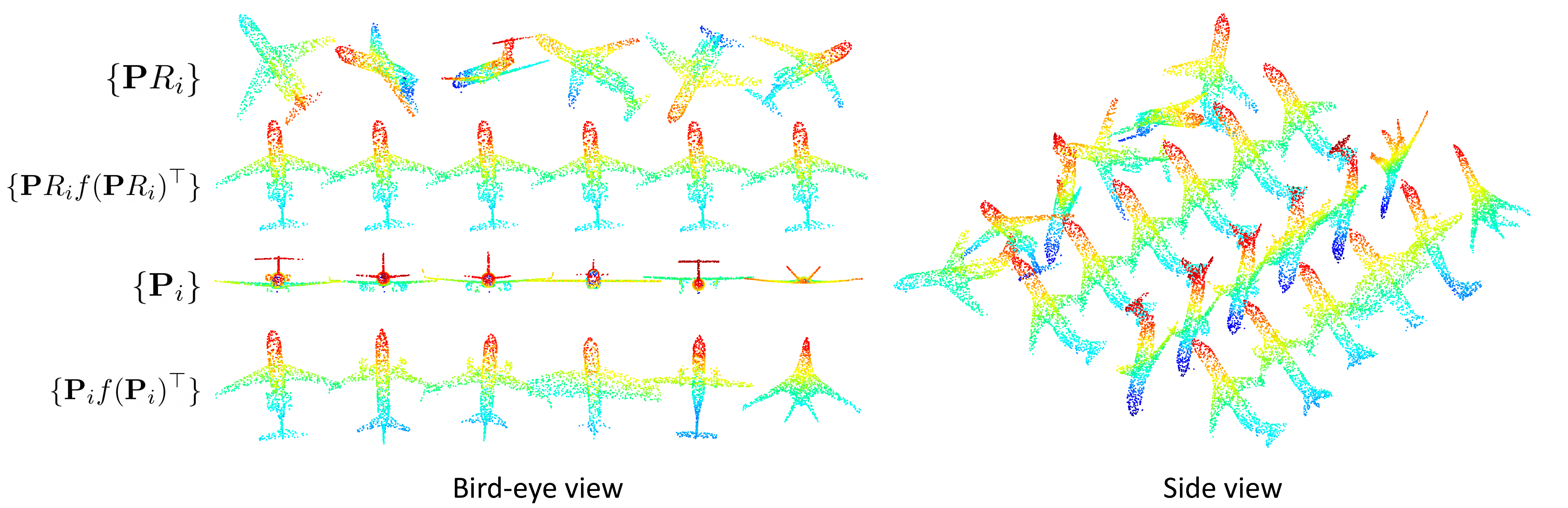}
    }
    \vspace{-7mm}
    \caption{
    \textbf{Qualitative visualization of canonicalized Airplane class on the ShapeNet dataset.}
    The first two rows visualize the stability of our characteristic orientation predictions.
    Given differently rotated point clouds of the same instance $\{\mathbf{P}R_i \}$, \methodName yields stable canonicalized point clouds $\{\mathbf{P}R_i f(\mathbf{P}R_i)^\top \}$ such that they are all well aligned.
    The lower two rows visualize the consistency of our characteristic orientation predictions.
    Given initially aligned different point clouds of the same class $\{\mathbf{P}_i\}$, \methodName is able to obtain consistent canoncalizations $\{\mathbf{P}_i f(\mathbf{P}_i )^\top \}$ such that they are aligned.
    Note that the pre-aligned coordinate frame of the dataset differs from the learned canonical orientation for we use a self-supervised loss $\mathcal{L}$ to learn to predict characteristic orientations.
    }
    \label{fig:stability_consistency}
\end{figure*}
\begin{table*}[!ht]
\caption{\textbf{Single-class stability ($^{\circ}$) and consistency ($^{\circ}$) comparison on the ShapeNet dataset.}
        Lower is better.
        While VN-SPD demonstrates the best stability, \methodName achieves the best consistency overall, and is just marginally behind VN-SPD in terms of stability.
        }
\centering
\resizebox{0.95\textwidth}{!}{%
\begin{tabular}{lcc|cc|cc|cc}
                        \toprule
                        \multirow{2}{*}{Method} & \multicolumn{2}{c}{Airplane} & \multicolumn{2}{c}{Car} & \multicolumn{2}{c}{Chair} & \multicolumn{2}{c}{Table} \\
                         & Stability & Consistency & Stability & Consistency & Stability & Consistency & Stability & Consistency \\
                         
                        \midrule
                        
                        Compass~\yrcite{spezialetti2020learning} & 13.81 & 71.43 & 12.01 & 68.20 & 19.20 & 87.50 & 74.80 & 115.3 \\
                        
                        Canonical Capsules~\yrcite{sun2021canonical} & 7.42 & \underline{45.76} & 4.79 & 68.13 & 81.9 & \textbf{11.1} & 14.7 & 119.3 \\

                         ConDor~\yrcite{sajnani2022condor} & 35.93 & 118.00 & 34.52 & 109.55 & 25.98 & 122.08 & 29.68 & \textbf{77.99} \\
                        
                        VN-SPD~\yrcite{katzir2022shape} & \textbf{0.02} & 49.97 & \textbf{0.04} & \underline{24.31}  & \textbf{0.02} & 35.6 & \textbf{0.02} & 106.3 \\
                        
                        \methodName (Ours) & \underline{0.67} & \textbf{32.07} & \underline{0.24} & \textbf{13.52} & \underline{0.33} & \underline{18.91} & \underline{0.85} & \underline{101.23} \\

                        \bottomrule
    \end{tabular}
}
\label{tbl:stability_consistency}
\end{table*}

\begin{table*}[!ht]
\caption{\textbf{Multi-class stability ($^{\circ}$) and consistency ($^{\circ}$) comparison on the ShapeNet dataset}
        Lower is better.
        It can be seen that \methodName demonstrates the best stability and consistency overall, outperforming existing methods by a significant margin especially in terms of stability.
        Point Transformer* is a SO(3)-\textbf{variant} characteristic orientation predictor based on Point Transformer~\cite{zhao2021point}, optimized in favor of consistency while neglecting stability; this model will be used to analyze the effect of stability and consistency in downstream point cloud analysis tasks.
        }
\centering
\resizebox{0.95\textwidth}{!}{%
\begin{tabular}{lcc|cc|cc|cc}
                        \toprule
                        \multirow{2}{*}{Method} & \multicolumn{2}{c}{Airplane} & \multicolumn{2}{c}{Car} & \multicolumn{2}{c}{Chair} & \multicolumn{2}{c}{Table} \\
                         & Stability & Consistency & Stability & Consistency & Stability & Consistency & Stability & Consistency \\
                         
                        \midrule
                        
                        Canonical Capsules~\yrcite{sun2021canonical} & \underline{20.96} & 129.03 & \underline{7.08} & \underline{78.29} & \underline{9.11} & 109.04 & \underline{18.62} & 123.25 \\

                        ConDor~\yrcite{sajnani2022condor} & 31.05 & 122.86 & 34.17 & 113.83 & 27.07 & 118.89 & 31.22 & 128.89 \\
                        
                        VN-SPD~\yrcite{katzir2022shape} & 97.88 & \underline{104.82} & 96.99 & 95.63 & 97.96 & \underline{94.91} & 97.19 & \textbf{97.72} \\
                        
                        \methodName (Ours) & \textbf{0.67} & \textbf{48.72} & \textbf{0.40} & \textbf{21.77} & \textbf{0.42} & \textbf{25.65} & \textbf{4.80} & \underline{103.64} \\
                        \midrule
                        Point Transformer*~\yrcite{zhao2021point} & 86.12 & 16.00 & 85.89 & 21.19 & 85.44 & 13.07 & 87.13 & 18.36 \\
                   
                        \bottomrule
    \end{tabular}
}
\label{tbl:stability_consistency_multi}
\end{table*}

\subsection{Characteristic Orientation Prediction}

\noindent
\textbf{Dataset and implementation details.}
We use the ShapeNet dataset~\cite{shapenet} to train and test our model for characteristic orientation prediction of point clouds. 
Following VN-SPD~\cite{katzir2022shape}, we focus on four classes of the ShapeNet dataset: airplanes, chairs, tables, and cars.
Note that airplanes and cars classes are more semantically consistent \ie, contain less intra-class shape variation, while chairs and tables are less semantically consistent.
We sample 1024 points randomly for each point cloud.
We train our \methodName network using the Adam optimizer~\cite{kingma2014adam} at a learning rate of 0.01 for 3000 epochs.
We use the checkpoint which shows the best validation sum of stability and consistency (lower the better) as our model for evaluation i.e., the model with the best compromise between stability and consistency on the validation set.
To facilitate the analysis of the influence of stability and consistency on downstream tasks, we also propose to train a model that places more focus on consistency than stability, which is referred to as Point Transformer*.
Point Transformer* is based on a Point Transformer network~\cite{zhao2021point} that takes as input the SO(3)-equivariant feature $\phi (\mathbf{P}R)$ from our hypothesizer network, and the rotated input point cloud $\mathbf{P}R$.
As the layers of Point Transformer violate SO(3)-equivariance, the stability of characteristic orientation predictions is neglected.
We use its output to predict the characteristic orientation of the input as well.

For each class, we calculate the stability for each instance in the validation set by applying 10 random rotations using Eq. \eqref{eq:stability} and report their mean.
In the case of consistency, we follow Eq. \eqref{eq:consistency} and report the consistency for each class. To enforce the uniqueness of the Singular Value Decomposition process necessary to calculate the chordal $L_2$-mean of rotations, we use the mean rotation whose determinant is 1.

\begin{figure*}[!t]
    \centering
    \resizebox{\textwidth}{!}{
    \includegraphics{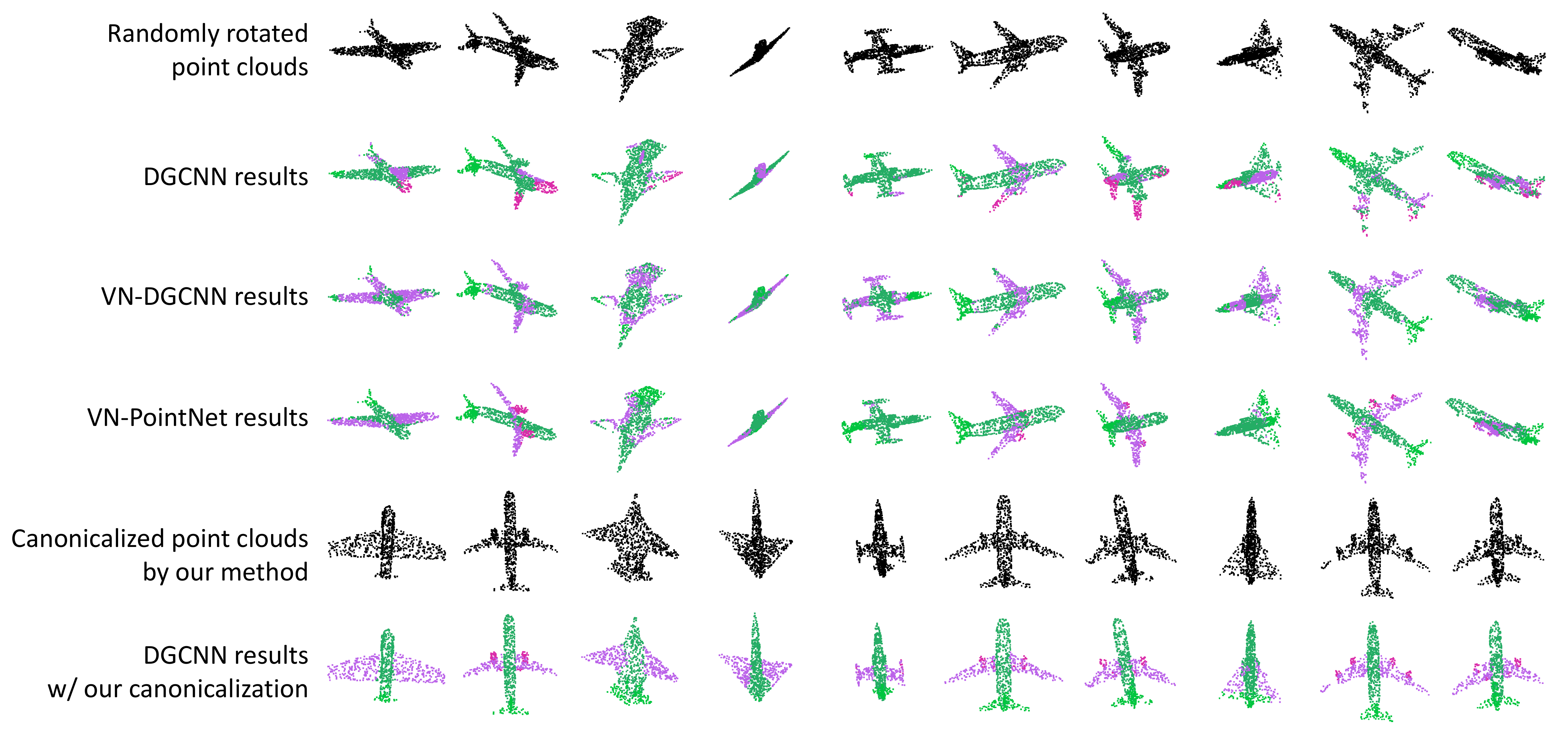}
    }
    \vspace{-6mm}
    \caption{
    \textbf{Qualitative comparison of point cloud part segmentation on the ShapeNet~\cite{shapenet} dataset.} 
    We use the I/SO(3) setting, where no rotation augmentation is applied during training, but the point clouds are randomly rotated at test time.
    The first two rows visualize vanilla DGCNN~\cite{dgcnn}, where it can be seen that the part segmentation results fail to differentiate the wings of the plane from the body.
    The next two rows visualize Vector Neuron-based networks~\cite{deng2021vector}, which can predict SO(3)-invariant point-wise part semantics. However, due to their limited representation power, these SO(3)-invariant networks still fail to discriminate detailed parts (\eg, wings) of airplanes.
    Finally, the last two rows visualize DGCNN using \methodName to canonicalize all input point clouds at train and test time.
    It can be seen that the input point clouds are well aligned, resulting in significantly better part segmentation results.
    }
    \label{fig:shapenet_part_seg}
\end{figure*}
\begin{table*}[!ht]
\caption{
\textbf{Class-wise mIoU (\%) and Instance-wise mIoU (\%) on ShapeNet for part segmentation.}
The higher the better.
We guide the readers to the main text for details on the I/I, I/SO(3) and SO(3)/SO(3) evaluation settings.
Vanilla DGCNN, which does not use any aligning method, performs the best at I/I setting, hinting that no aligning method is perfectly stable \textit{and} consistent.
However, at I/SO(3) and SO(3)/SO(3) settings, \methodName outperforms other methods.
Meanwhile, Vector Neuron-based networks including VN-PointNet are all outperformed by DGCNN with aligning networks due to their limited representation power.
Note that Point Transformer* achieved the best multi-class consistency, while showing substandard stability.
This evidences that having an optimal balance between stability and consistency is the key to yielding better results on downstream tasks, rather than sacrificing stability for consistency.
}
\centering
\resizebox{0.83\linewidth}{!}{%
\begin{tabular}{lcc|cc|cc}
    \toprule
    \multirow{2}{*}{Method} & \multicolumn{2}{c}{I/I} & \multicolumn{2}{c}{I/SO(3)} & \multicolumn{2}{c}{SO(3)/SO(3)}  \\
     & Cls. mIoU & Ins. mIoU & Cls. mIoU & Ins. mIoU & Cls. mIoU & Ins. mIoU \\

    \midrule

    VN-PointNet~\yrcite{deng2021vector} & 58.16 & 66.99 & 58.16 & 66.99 & 58.22 & 67.37 \\
    VN-DGCNN~\yrcite{deng2021vector} & 52.72 & 63.18 & 52.72 & 63.18 & 52.52 & 62.85 \\
     
    \midrule
    
    DGCNN~\yrcite{dgcnn} & \textbf{77.27} & \textbf{81.66} & 25.81 & 29.06 & 58.59 & 69.09 \\
    ~~~~+ Point Transformer*~\yrcite{zhao2021point} & 66.25 & \underline{76.95} & 43.44 & 51.41 & 62.95 & 73.15 \\
    ~~~~+ Canonical Capsules~\yrcite{sun2021canonical} & 64.02 & 74.32 & \underline{63.92} & \underline{74.23} & \underline{64.45} & \underline{74.69} \\
    ~~~~+ ConDor~\yrcite{sajnani2022condor} & 59.00 & 69.71 & 59.07 & 69.81 & 58.87 & 69.75 \\
    ~~~~+ VN-SPD~\yrcite{katzir2022shape} & 63.29 & 73.18 & 48.25 & 57.11 & 59.84 & 69.90 \\
    ~~~~+ \methodName (Ours) & \underline{68.84} & 76.64 & \textbf{68.88} & \textbf{76.67} & \textbf{68.12} & \textbf{76.46} \\    

    \bottomrule
    \end{tabular}
}
\label{tbl:shapenet_segmentation}
\end{table*}

\noindent
\textbf{Results and analyses.} For each method, we compare two types of models for this experiment: class-wise models and multi-class models. 
The class-wise models are trained on a single class of shapes, while multi-class models are trained on the union of the four ShapeNet classes used in our experiments, facilitating the comparison of multi-class training suitability between different methods.
\Tbl{stability_consistency} shows the results of single-class models, and \Tbl{stability_consistency_multi} presents those of multi-class models.

\begin{table*}[!ht]
\caption{\textbf{Stability ($^{\circ}$) and consistency ($^{\circ}$) results of ablation study on the ShapeNet~\cite{shapenet} airplane class.}
        Starting from VN-SPD~\cite{katzir2022shape} as the baseline, we conduct an ablation study to verify the effect of each component of \methodName.
        It can be seen that while removing the reconstruction pipeline has negative effects on both stability and consistency, finally incorporating our cross-instance training and residual predictor yields significantly better consistency results for a negligible decline in stability.
        Note that we empirically showed that removing the reconstruction pipeline makes our architecture (and the Point Transformer* architecture) more suitable to be trained using multiple classes of point clouds.
        }
\centering
\resizebox{0.85\textwidth}{!}{%
\begin{tabular}{l|ccc|cc}
                        \toprule

                        Method & w/o Recon. & Cross-instance & Residual Predictor $g(\cdot)$ & Stability & Consistency \\
                         
                        \midrule

                        VN-SPD & & & & 0.02 & 49.97 \\
                        \methodName w/o cross-instance and $g(\cdot)$ & \checkmark & & & 0.49 & 61.06 \\
                        \methodName w/o $g(\cdot)$ & \checkmark & \checkmark & & 0.68 & 36.25 \\
                        \methodName (Ours) & \checkmark & \checkmark & \checkmark & 0.67 & 32.07 \\
                   
                        \bottomrule
    \end{tabular}
}
\label{tbl:stability_consistency_ablation}
\end{table*}

For single-class model comparison, it can be seen that \methodName exhibits the best consistency overall - significantly outperforming baseline methods while being only slightly underperforming VN-SPD~\cite{katzir2022shape} on the stability metric.
In the case of multi-class model comparison, our model achieves the best stability and consistency compared to existing methods. %
This shows that using point reconstruction pipelines makes it unsuitable to handle multiple classes within a single model, in contrast to ours which is more adept at training with multiple classes of shapes.
However, \methodName still predicts characteristic orientations with the best balance between stability and consistency.

\subsection{Part Segmentation}
\label{exp_segmentation}
\noindent
\textbf{Dataset and implementation details.}
To analyze the influence of stability and consistency on downstream analysis tasks, we evaluate our method on the ShapeNet part segmentation dataset~\cite{shapenet}, specifically on the four data classes of airplane, chair, car, and table.
We first pre-train the characteristic orientation prediction methods, including \methodName, on the ShapeNet dataset under the multi-class training setting, then freeze their parameters.
For each method, we then train a DGCNN~\cite{dgcnn} network for part segmentation, where its input point clouds are canonicalized using the predicted characteristic orientation by each characteristic orientation prediction method, at both training and testing.
Following VNNs~\cite{deng2021vector}, we train the DGCNN network with Stochastic Gradient Descent optimizer at a learning rate of 0.001 for 200 epochs for all part segmentation experiments.
We evaluate \methodName against the baseline methods under three evaluation settings: I/I, I/SO(3), and SO(3)/SO(3). 
I/I provides no rotation to the input point cloud at both train and test time, such that all input point clouds are readily aligned at both train and test times.
I/SO(3) applies random rotation to the point cloud only at test time, such that the input point cloud would be randomly oriented at test time. 
SO(3)/SO(3) applies random rotation to the point cloud at both train and test time, such that all input point clouds are randomly rotated.
Note that applying random rotation to the point cloud at train time is the same as using rotation augmentations, but applying random rotations to the point cloud at test time differs from test-time augmentation as the former refers to viewing a single randomly rotated point cloud instance, while the latter refers to viewing multiple randomly rotated instances of the same input point cloud.

\noindent
\textbf{Results and analyses.}
We report the class-wise and instance-wise mIoU in \Tbl{shapenet_segmentation}.
At the I/I setting, because the input point clouds are already aligned, vanilla DGCNN yields the best results - however, among methods that do use aligning methods, using \methodName yields the best results.
At the I/SO(3) setting, vanilla DGCNN shows extremely poor results, because the DGCNN network has not learned to segment rotated point clouds. 
\methodName shows the best results in this setting, evidencing that canonicalizing the point clouds using our predicted characteristic orientation is the most beneficial.
At the SO(3)/SO(3) settings, vanilla DGCNN shows an improved performance in comparison to the I/SO(3) setting, but underperforms in comparison to using aligning, proving that rotation augmentations are insufficient to handle randomly rotated point clouds at test time. 
\methodName also achieves the best performance in this setting.
Noting that Point Transformer* showed a significantly higher consistency compared to \methodName, but \methodName exhibits the best performance overall, our takeaway is that finding a harmonious balance between stability and consistency of characteristic orientation predictions is crucial for downstream analysis tasks.

\begin{table*}[t!]
\begin{minipage}{0.42\linewidth}
\centering
\caption{\textbf{The numerical error of $k$NN indices.} We count the number of wrong $k$NN edges of rotated point clouds on ShapeNet~\cite{shapenet} test dataset under two different precision levels. We set the cardinality of each point cloud as 1024 and $k$ as 20, which means 20480 edges in total.}
\scalebox{0.85}{
\begin{tabular}{c|c|c|c|c}
        \toprule
        Dtype & Airplane & Car & Chair & Table \\
        \midrule
        Float32 & 17.4 & 9.6 & 27.9 & 62.3 \\
        Float64 & 15.7 & 7.8 & 25.8 & 58.2 \\
        \bottomrule
        \end{tabular}
}
\label{tbl:floating_point_precision_KNN_error}
\end{minipage}
\hfill
\begin{minipage}{0.55\linewidth}
\centering
\caption{\textbf{On the training pair and $k$NN fixes (single class).} We conduct the experiment on ShapeNet~\cite{shapenet} airplane class.}
\scalebox{0.85}{
\begin{tabular}{l|c|c}
        \toprule
        Method & Stability & Consistency \\
        \midrule
        VN-SPD~\yrcite{katzir2022shape} & \textbf{0.02} & 49.97 \\
        \methodName (Ours) & 0.67 & \textbf{32.07} \\
        \methodName w/ fixed $k$NN & \textbf{0.02} & \textbf{32.07} \\
        \methodName w/ same-instance & 0.14 & \underline{47.08} \\
        \methodName w/ same-instance \& fixed $k$NN & \underline{0.03} & \underline{47.08} \\
        \bottomrule
        \end{tabular}
}
\label{tbl:single_after_fixes}
\end{minipage}
\end{table*}
\begin{table*}[!t]
\vspace{-2mm}
\caption{\textbf{On the training pair and $k$NN fixes (multi classes).} Similar to the result of single class experiment~\Tbl{single_after_fixes}, ours shows a better stability with the same-instance pair and $k$NN fixes. We denote our \methodName with the same-instance pair as \methodName*, \methodName with fixed $k$NN as \methodName$^\dagger$, and \methodName with both modifications as \methodName*$^\dagger$, respectively.}
\centering
\resizebox{0.9\textwidth}{!}{%
\begin{tabular}{lcc|cc|cc|cc}
                        \toprule
                        \multirow{2}{*}{Method} & \multicolumn{2}{c}{Airplane} & \multicolumn{2}{c}{Car} & \multicolumn{2}{c}{Chair} & \multicolumn{2}{c}{Table} \\
                         & Stability & Consistency & Stability & Consistency & Stability & Consistency & Stability & Consistency \\
                         
                        \midrule
                        
                        VN-SPD~\yrcite{katzir2022shape} & 97.88 & 104.82 & 96.99 & 95.63 & 97.96 & 94.91 & 97.19 & \textbf{97.72} \\
                        
                        \methodName (Ours) & 0.67 & \textbf{48.72} & \underline{0.40} & \textbf{21.77} & 0.42 & \textbf{25.65} & 4.80 & \underline{103.64} \\

                        \methodName$^\dagger$ & \textbf{0.02} & \textbf{48.72} & \textbf{0.02} & \textbf{21.77} & \textbf{0.02} & \textbf{25.65} & \underline{0.03} & \underline{103.64} \\

                        \methodName* & \underline{0.04} & \underline{56.63} & \textbf{0.02} & \underline{26.76} & \underline{0.03} & \underline{35.87} & 0.23 & 107.10 \\

                        \methodName*$^\dagger$ & \textbf{0.02} & \underline{56.63} & \textbf{0.02} & \underline{26.76} & \textbf{0.02} & \underline{35.87} & \textbf{0.02} & 107.10 \\
                   
                        \bottomrule
    \end{tabular}
}
\label{tbl:multi_after_fixes}
\end{table*}

\subsection{Ablation Study and Analyses}
\label{ablation}

\noindent
\textbf{Ablation study.}
We perform an ablation study to demonstrate the efficacy of each technique and component in  \methodName, starting from VN-SPD~\cite{katzir2022shape} as the baseline, given that they share the same SE(3)-equivariant VNN~\cite{deng2021vector}-based backbone network as \methodName.
The results are illustrated in \Tbl{stability_consistency_ablation}.

First, it is evident that removing the point cloud reconstruction pipeline~\cite{groueix2018papier} from VN-SPD deteriorates both the stability and consistency, which shows that the disentanglement of pose and shape by leveraging point cloud reconstruction is indeed helpful.
Next, we verify that using cross-instance training \ie, using point clouds pairs with different point clouds from the same class significantly improves the performance, already outperforming VN-SPD by a large margin of approximately $14^{\circ}$.
Finally, we demonstrate that incorporating our residual predictor, $g(\cdot)$, improves the consistency even further, without harming the stability of characteristic orientation predictions.

\noindent
\textbf{Further analysis on stability of \methodName.}
While \methodName strikes a much favorable balance between stablity and consistency compared to VN-SPD, VN-SPD shows slightly favorable stability by an average value of $0.5^\circ$.
Through further analyses, we determined that the floating point precision loss of 3D point coordinates causes the results of $k$NN on point clouds to differ before and after rotation (illustrated in \Tbl{floating_point_precision_KNN_error}), which leads to slight instabilities in the SO(3)-equivariant calculations of our network.

Another factor that affects the stability of \methodName can be deduced from \Tbl{stability_consistency_ablation}, where it can be seen that the usage of our residual predictor and cross-shape dataset pairs incurs slight declines in stability in return for significant gains in consistency.
Instead of using cross-shape dataset pairs, we could train \methodName using same-shape pairs with augmentation transformations to enforce consistency of characteristic orientation predictions while preserving stability.

We present the results of \methodName in \Tbl{single_after_fixes} for the single-class setting and \Tbl{multi_after_fixes} for the multi-class setting, when we fix the $k$NN indices for each point cloud beforehand at inference time, and when we use same-shape dataset pairs to train \methodName.
It can be seen that using fixed $k$NN indices results in state-of-the-art stability, demonstrating nearly zero stability even under the multi-class setting.
Albeit VN-SPD does not use predetermined $k$NN indices, their training scheme focuses on same-shape pair training with stronger data augmentations, resulting in satisfactory stability under the single-class setting.
Also, using same-shape pairs results in state-of-the-art stability in both single- and multi-class settings, with a slight increase in consistency in comparison to using cross-shape dataset pairs.
While it would be favorable to be able to have consistent $k$NN predictions, this is difficult to facilitate in real-world scenarios; it would therefore be wiser to integrate same-shape dataset pairs together with cross-shape pairs to strike the best balance between stability and consistency of characteristic orientation predictions.

\section{Conclusion}
\label{sec:conclusion}

We have shown that \methodName learns to assign a stable yet consistent characteristic orientation for an input point cloud.
It leverages SO(3)-equivariant backbone networks to determine the characteristic orientation hypothesis of the input point cloud, and calibrates the hypothesis with the results of our novel residual predictor to successfully canonicalize the input point cloud.
We have evidenced the superior stability and consistency of \methodName in both quantitative and qualitative manners.
Specifically, the experiment results on characteristic orientation prediction indicate that by removing the conventional point cloud reconstruction pipelines, \methodName is more suitable to be trained using multiple classes of point clouds, being more effective for downstream tasks \eg, point cloud part segmentation.
Furthermore, by comparing different models with varying balances between stability and consistency on downstream tasks, we draw an insightful conclusion that the key to improving the performance of downstream tasks is an optimal compromise between stability and consistency.

{
\small
\noindent\textbf{Acknowledgement.}
This work was supported by the MX division of Samsung Electronics Co., Ltd., the Institute of Information \& Communications Technology Planning \& Evaluation (IITP) grants (No.2021-0-02068: AI Innovation Hub (50\%), No.2022-0-00290: Visual Intelligence based on Multi-layered Visual Common Sense (40\%), No.2019-0-01906: AI Graduate School Program at POSTECH (10\%)) funded by Korea government (MSIT).
}

\clearpage
\bibliography{references}
\bibliographystyle{icml2023}

\newpage
\appendix
\onecolumn

\section{On the Capacity of \methodName}

In principle, using a large enough network with a strong enough supervision from a \textit{large-scale} dataset should be competitive to our choice of leveraging SO(3)-equivariant networks which maintain the inductive bias on rotation while consequently incurring capacity limits.
Using a large enough network to rely on strong enough supervision to learn equivariance has two disadvantages: 1) while it has to solely rely on a sufficiently large-scale dataset, it severely lacks sample efficiency, and still would not guarantee the generalization of equivariance, and 2) this poses difficulties in guaranteeing stability while aiming to improve consistency of predictions. 
On the other hand, our design of predicting the SO(3)-invariant residual while maintaining SO(3)-equivariance has two advantages: 1) our method is theoretically guaranteed to satisfy SO(3)-equivariance, and 2) we can therefore exploit sample efficiency in training.
The main reason we maintain the inductive bias on rotation in \methodName is for its high sampling efficiency, noting that we do not need to train our model with large-scale datasets with extensive rotation augmentations to learn equivariance to rotation.

To provide deeper insights into this matter, we train two different straightforward approaches - 1) the vanilla Point Transformer, and 2) our Point Transformer* architecture, which is Point Transformer followed by our SO(3)-equivariant hypothesizer - with varying Point Transformer sizes in terms of parameter. 
Note that both of these approaches break SO(3)-equivariance. Specifically, we newly train a total of six networks for the task of characteristic orientation prediction:

\begin{itemize}
  \item PT-S: vanilla Point Transformer with 4 transformer blocks (the same number of blocks as \methodName)
  \item PT-M: vanilla Point Transformer with 8 transformer blocks
  \item PT-L: vanilla Point Transformer with 16 transformer blocks
  \item PT*-S: SO(3)-equivariant hypothesizer + PT-S
  \item PT*-M: SO(3)-equivariant hypothesizer + PT-M
  \item PT*-L: SO(3)-equivariant hypothesizer + PT-L
\end{itemize}

Note that PT-variants tend to diverge when trained with the same learning rate as \methodName, and were trained with 1/100 times the learning rate. 
For a fair comparison, we additionally train PT*- variants with this new learning rate (1/100 of that of \methodName), which we name PT*-lr-S/M/L.
The results are illustrated in \Tbl{point_transformer_variants}.

\begin{table*}[!h]
\caption{\textbf{Multi-class stability ($^\circ$) and consistency ($^\circ$) using SO(3)-variant backbones with varying capacity on ShapeNet.} 
The results do not show a clear correlation between stability/consistency and the model capacity.
\methodName shows a stronger characteristic orientation prediction performance in comparison.
Note that VN-SPD has a SO(3)-equivariant encoder (341K parameters) and a reconstruction network (17,069K parameters).}
\centering
\resizebox{0.9\textwidth}{!}{%
\begin{tabular}{lr|cc|cc|cc|cc}
                        \toprule
                        \multirow{2}{*}{Method} & \multirow{2}{*}{\#Param(K)} & \multicolumn{2}{c}{Airplane} & \multicolumn{2}{c}{Car} & \multicolumn{2}{c}{Chair} & \multicolumn{2}{c}{Table} \\
                         & & Stability & Consistency & Stability & Consistency & Stability & Consistency & Stability & Consistency \\
                         
                        \midrule
                        
                        VN-SPD~\yrcite{katzir2022shape} & 17410 & 97.88 & 104.82 & 96.99 & 95.63 & 97.96 & 94.91 & 97.19 & {97.72} \\
                        
                        \methodName (Ours) & 314 & \textbf{0.67} & {48.72} & \textbf{0.40} & {21.77} & \textbf{0.42} & {25.65} & \textbf{4.80} & 103.64 \\

                        PT-S &	\textbf{92}	&40.27&	58.83&	27.23&	42.69&	50.96&	66.97&	76.61&	75.42 \\
                        PT-M&	178 &	28.25 &	64.79 &	19.02 &	43.52	&19.16	&43.52&	65.59&	86.31\\
                        PT-L	&352	&33.42&	67.06&	21.04&	39.71	&24.43	&62.01	&70.69	&75.91\\
                        PT*-S&	135&	86.12&	\textbf{16.00}&	85.89&	21.19	&85.44&	\textbf{13.07}&	87.13&	\textbf{18.36}\\
                        PT*-M	&222	&85.34&	22.76&	80.67&	23.08	&86.12&	14.85	&81.38&	22.20\\
                        PT*-L	&396	&88.43&	35.61&	86.38&	\textbf{12.05}&	86.89&	20.62&	86.60&	26.13\\
                        PT*-lr-S&	135&	85.38&	52.63&	86.63&	28.17&	87.70&	33.07&	86.45	&35.60\\
                        PT*-lr-M&	222	&11.25	&62.64	&7.69&	32.73	&8.34	&40.49&	42.19&	83.29\\
                        PT*-lr-L&	396&	85.65&	36.15&	84.71&	24.87&	84.86&	45.07&	84.81&	52.25\\

                        \bottomrule
    \end{tabular}
}
\label{tbl:point_transformer_variants}
\end{table*}

The results show that the stability or consistency of characteristic orientation predictions does not correlate with the model capacity (\# parameters), and the point transformer variants overall show a much worse stability-consistency tradeoff in comparison with \methodName.
We suspect that this is mainly because the dataset used is not large enough to learn both the stability and consistency better than \methodName, which exploits the sampling efficiency of SO(3)-equivariant networks. 
An interesting observation is seen in PT*-lr-M, which actually demonstrates rather satisfactory stability compared to other PT- or PT*- variants, but they tend to show worse consistency in return. Overall, as of now, the capacity limits enforced by the usage of SO(3)-equivariance yields better results compared to models with larger learning capacities, which we mainly suspect is due to the lack of large-scale datasets (and conversely the sampling efficiency of SO(3)-equivariant networks).

\section{Effect of Using Test-time Augmentation}

We present additional quantitative results when applying test-time augmentation for the task of characteristic orientation prediction. 
Specifically, we use point resampling and Gaussian noise data augmentations for data augmentation at test time. 
The results are shown in \Tbl{tta_resampling} and \Tbl{tta_gaussian}, where it can be seen that applying resampling or noise augmentations at test time leads to worse results overall.
Nonetheless, \methodName still achieves the best stability-consistency tradeoff overall. 

\begin{table*}[!ht]
\caption{\textbf{Multi-class stability ($^{\circ}$) and consistency ($^{\circ}$) comparison on the ShapeNet dataset using resampling augmentation at test time.}
        Lower is better.
        It can be seen that \methodName strikes the most favorable balance between stability and consistency overall.
        }
\centering
\resizebox{0.95\textwidth}{!}{%
\begin{tabular}{lcc|cc|cc|cc}
                        \toprule
                        \multirow{2}{*}{Method} & \multicolumn{2}{c}{Airplane} & \multicolumn{2}{c}{Car} & \multicolumn{2}{c}{Chair} & \multicolumn{2}{c}{Table} \\
                         & Stability & Consistency & Stability & Consistency & Stability & Consistency & Stability & Consistency \\
                         
                        \midrule
                        
                        Canonical Capsules~\yrcite{sun2021canonical} & 76.26&	129.03&	57.16&	78.29&	46.32&	109.04&	85.40&	123.25 \\
                        
                        ConDor~\yrcite{sajnani2022condor} & 98.87&	122.86&	92.44&	113.83&	87.01&	118.89&	98.38&	128.89 \\

                        VN-SPD~\yrcite{katzir2022shape} & 100.21	&104.82&	100.87&	95.63&	99.98&	94.91&	101.25&	97.72 \\
                        
                        \methodName (Ours) & \textbf{13.53}	&48.72&	\textbf{7.69}&	21.77	&\textbf{7.56}&	25.65&	\textbf{66.67}	&103.64 \\
                        \midrule
                        Point Transformer*~\yrcite{zhao2021point} & 86.38&	\textbf{16.00}	&86.08&	\textbf{21.19}&	85.72	&\textbf{13.07}	&87.56	&\textbf{18.36} \\
                   
                        \bottomrule
    \end{tabular}
}
\label{tbl:tta_resampling}
\end{table*}
\begin{table*}[!ht]
\caption{\textbf{Multi-class stability ($^{\circ}$) and consistency ($^{\circ}$) comparison on the ShapeNet dataset using Gaussian noise augmentation at test time.}
        Lower is better.
        It can be seen that \methodName strikes the most favorable balance between stability and consistency overall.
        }
\centering
\resizebox{0.95\textwidth}{!}{%
\begin{tabular}{lcc|cc|cc|cc}
                        \toprule
                        \multirow{2}{*}{Method} & \multicolumn{2}{c}{Airplane} & \multicolumn{2}{c}{Car} & \multicolumn{2}{c}{Chair} & \multicolumn{2}{c}{Table} \\
                         & Stability & Consistency & Stability & Consistency & Stability & Consistency & Stability & Consistency \\
                         
                        \midrule
                        
                        Canonical Capsules~\yrcite{sun2021canonical} & 22.81&	129.03&	9.23&	78.29&	10.56&	109.04&	21.52&	123.25 \\

                        ConDor~\yrcite{sajnani2022condor} & 58.39&	122.86&	50.89&	113.83&	46.90&	118.89&	54.61&	128.89 \\
                        
                        VN-SPD~\yrcite{katzir2022shape} & 98.42	&104.82&	97.31&	95.63&	98.41&	94.91&	98.51&	97.72 \\
                        
                        \methodName (Ours) & \textbf{12.00}&	48.72&	\textbf{5.91}	&21.77&	\textbf{5.34}	&25.65&	45.30	&103.64 \\
                        \midrule
                        Point Transformer*~\yrcite{zhao2021point} & 86.88	&\textbf{16.00}	&86.09&	\textbf{21.19}	&85.73	&\textbf{13.07}	&87.34	&\textbf{18.36} \\
                   
                        \bottomrule
    \end{tabular}
}
\label{tbl:tta_gaussian}
\end{table*}
\clearpage
\section{Additional Qualitative Results}
\begin{figure*}[!h]
    \centering
    \resizebox{0.95\textwidth}{!}{
    \includegraphics{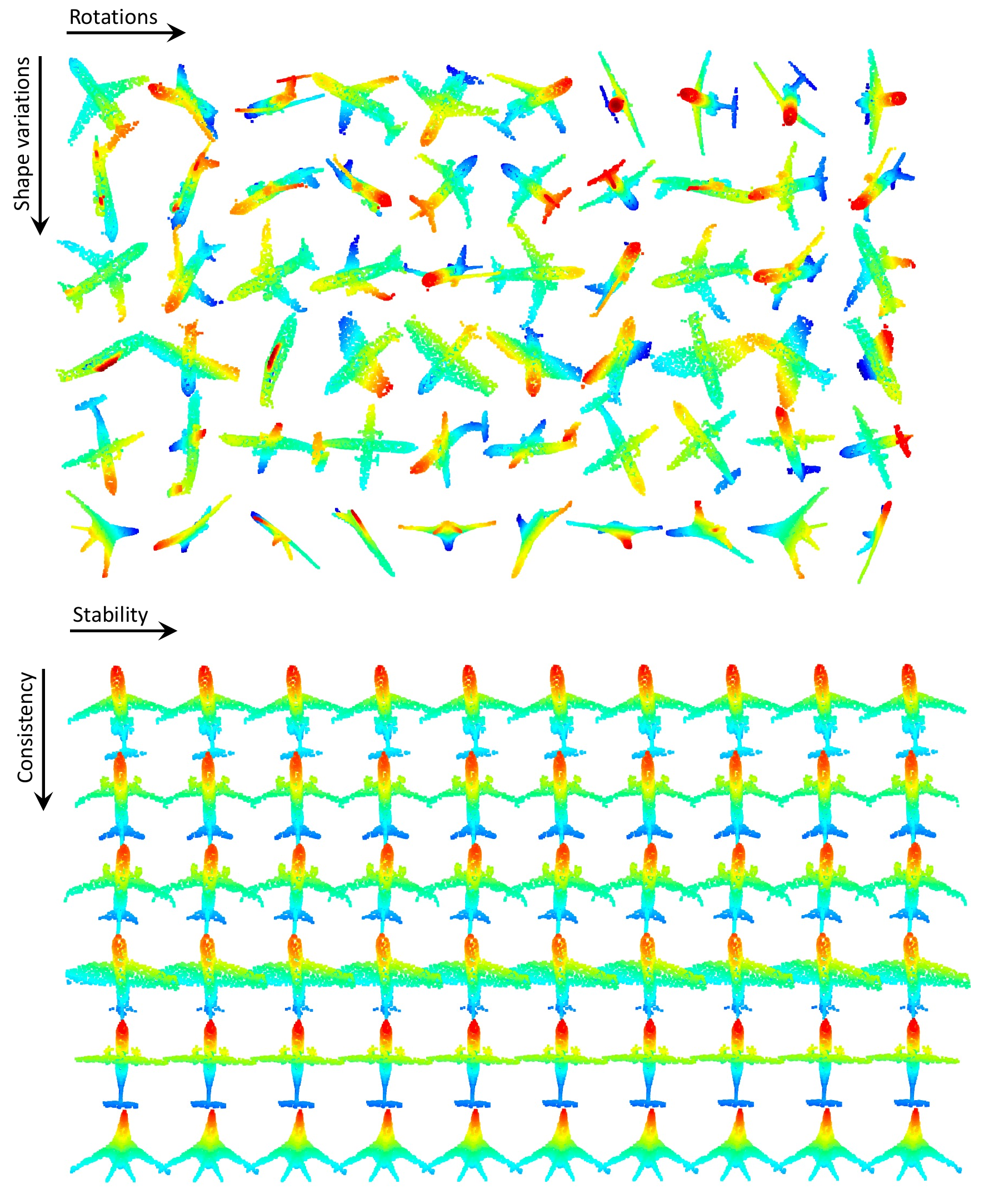}
    }
    \caption{\textbf{Qualitative visualization of canonicalized airplanes by our method (single-class) on the ShapeNet dataset.}
    }
    \label{fig:supp_plane}
\end{figure*}

\begin{figure*}[!t]
    \centering
    \resizebox{\textwidth}{!}{
    \includegraphics{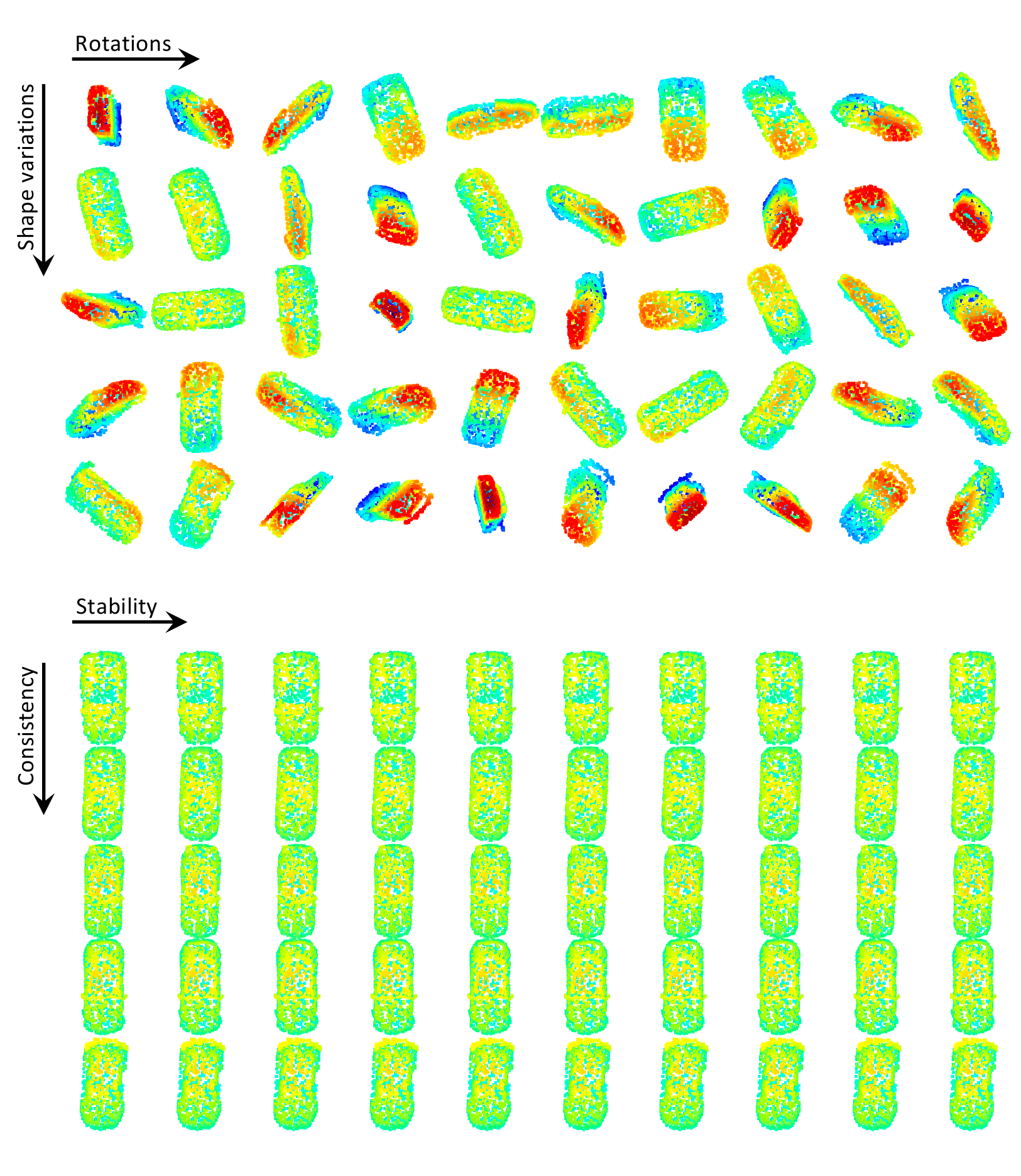}
    }
    \caption{\textbf{Qualitative visualization of canonicalized cars by our method (single-class) on the ShapeNet dataset.}
    }
    \label{fig:supp_car}
\end{figure*}

\begin{figure*}[!t]
    \centering
    \resizebox{0.9\textwidth}{!}{
    \includegraphics{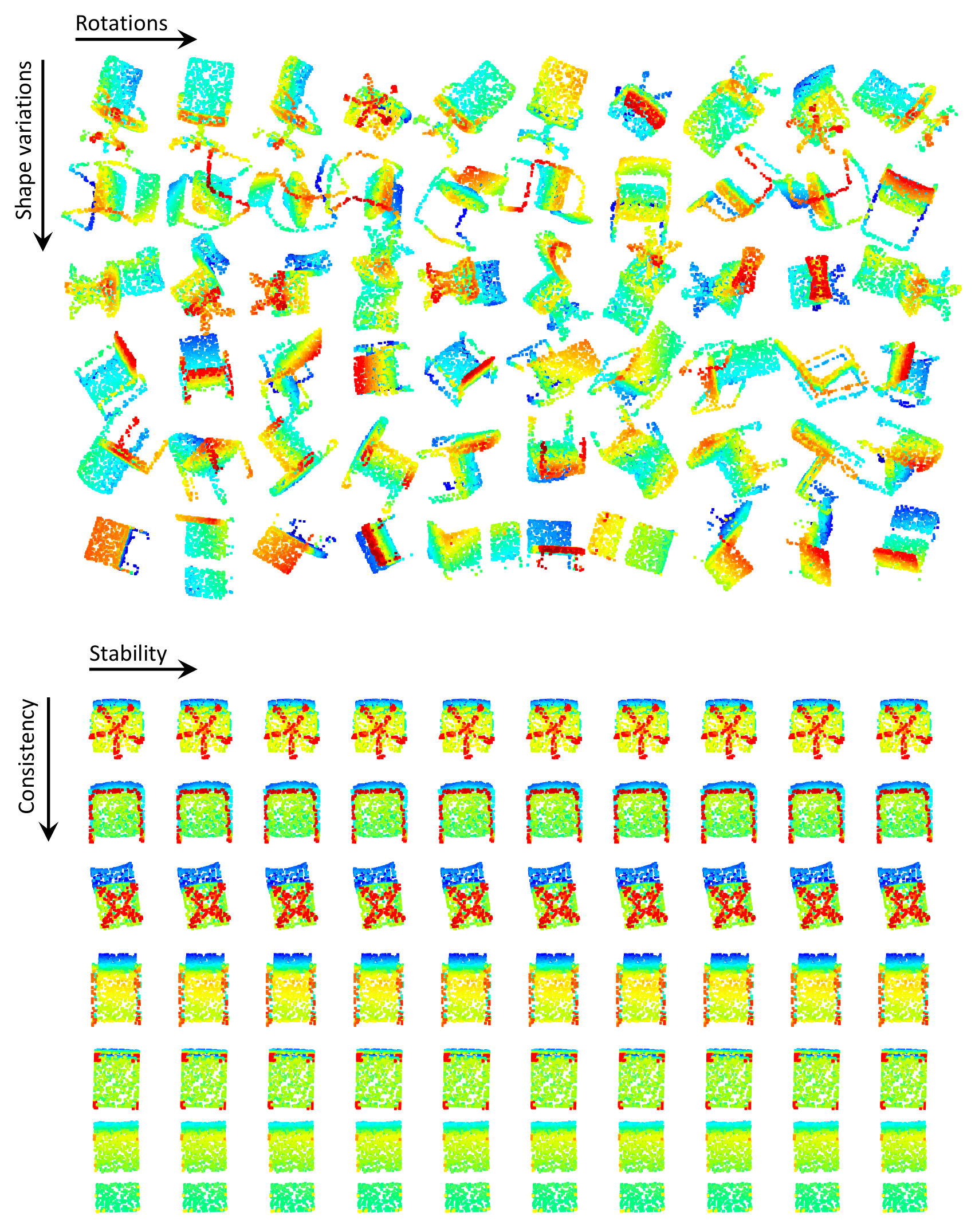}
    }
    \caption{\textbf{Qualitative visualization of canonicalized chairs by our method (single-class) on the ShapeNet dataset.}
    }
    \label{fig:supp_chair}
\end{figure*}

\begin{figure*}[!h]
    \centering
    \resizebox{0.85\textwidth}{!}{
    \includegraphics{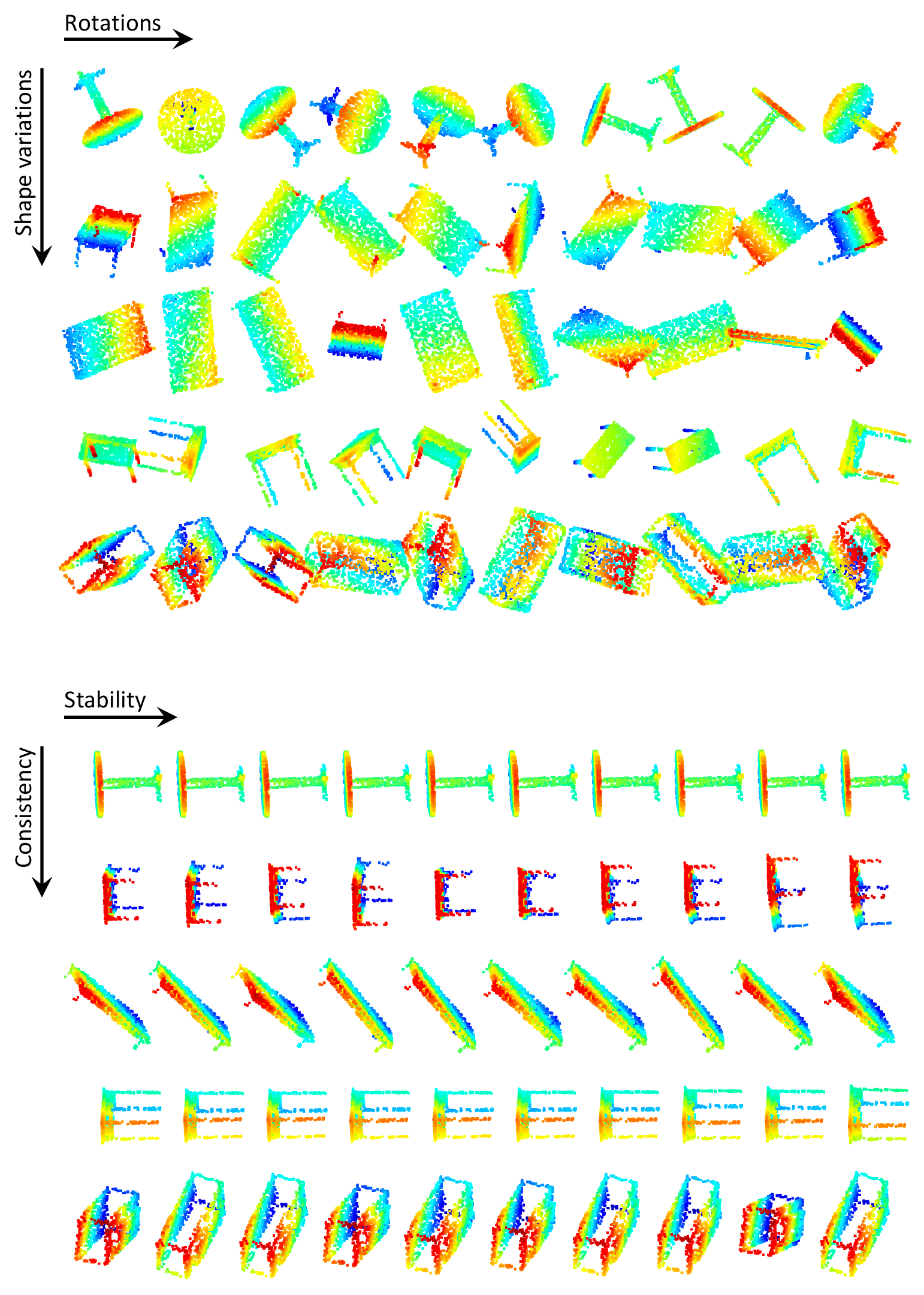}
    }
    \caption{\textbf{Qualitative visualization of canonicalized tables by our method (single-class) on the ShapeNet dataset.}
    }
    \label{fig:supp_table}
\end{figure*}

\begin{figure*}[!t]
    \centering
    \resizebox{\textwidth}{!}{
    \includegraphics{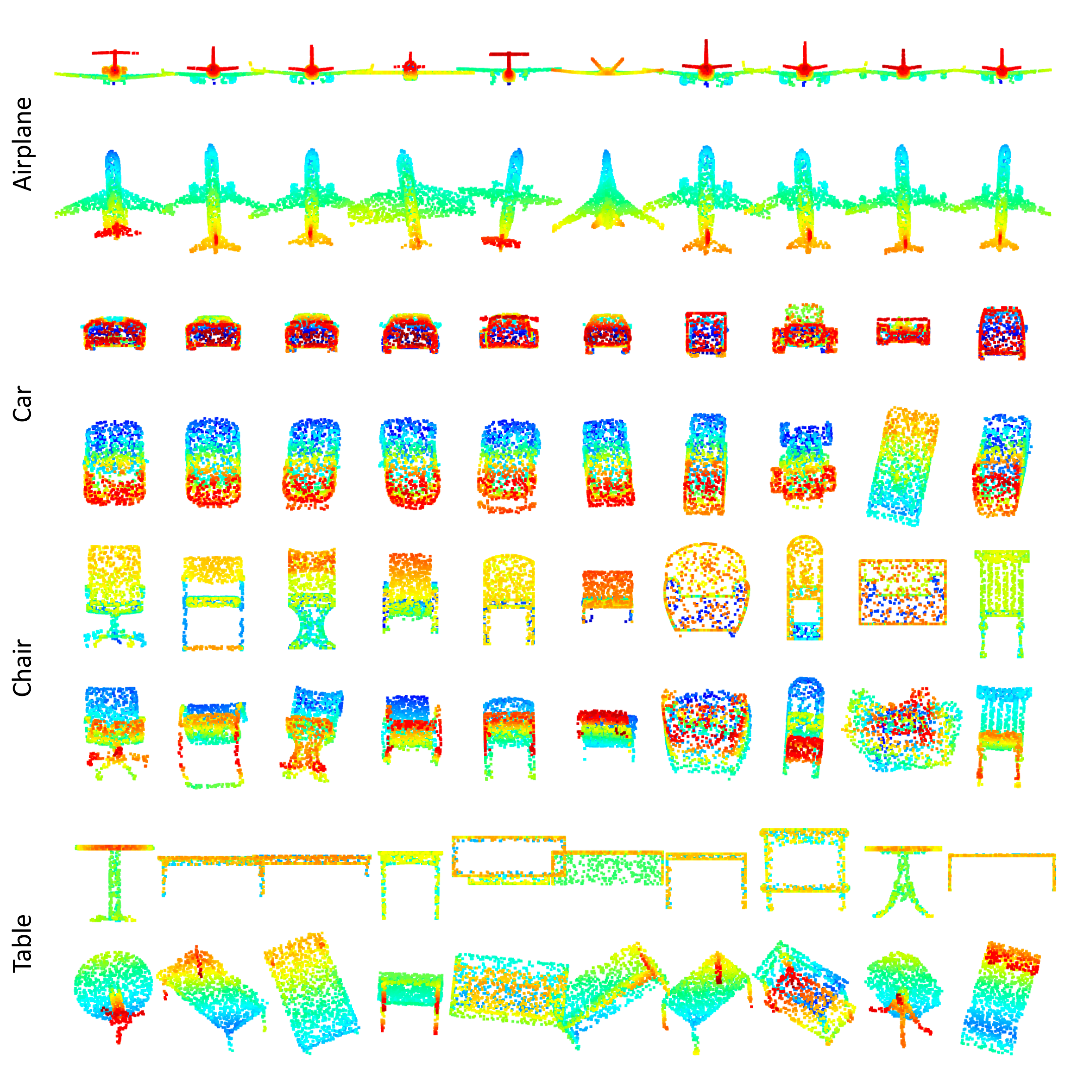}
    }
    \caption{\textbf{Qualitative visualization of canonicalized point clouds of four classes.} All point clouds are canonicalized by our method, which is trained with multi-class training.
    }
    \label{fig:supp_multi}
\end{figure*}

\begin{figure*}[!t]
    \centering
    \resizebox{\textwidth}{!}{
    \includegraphics{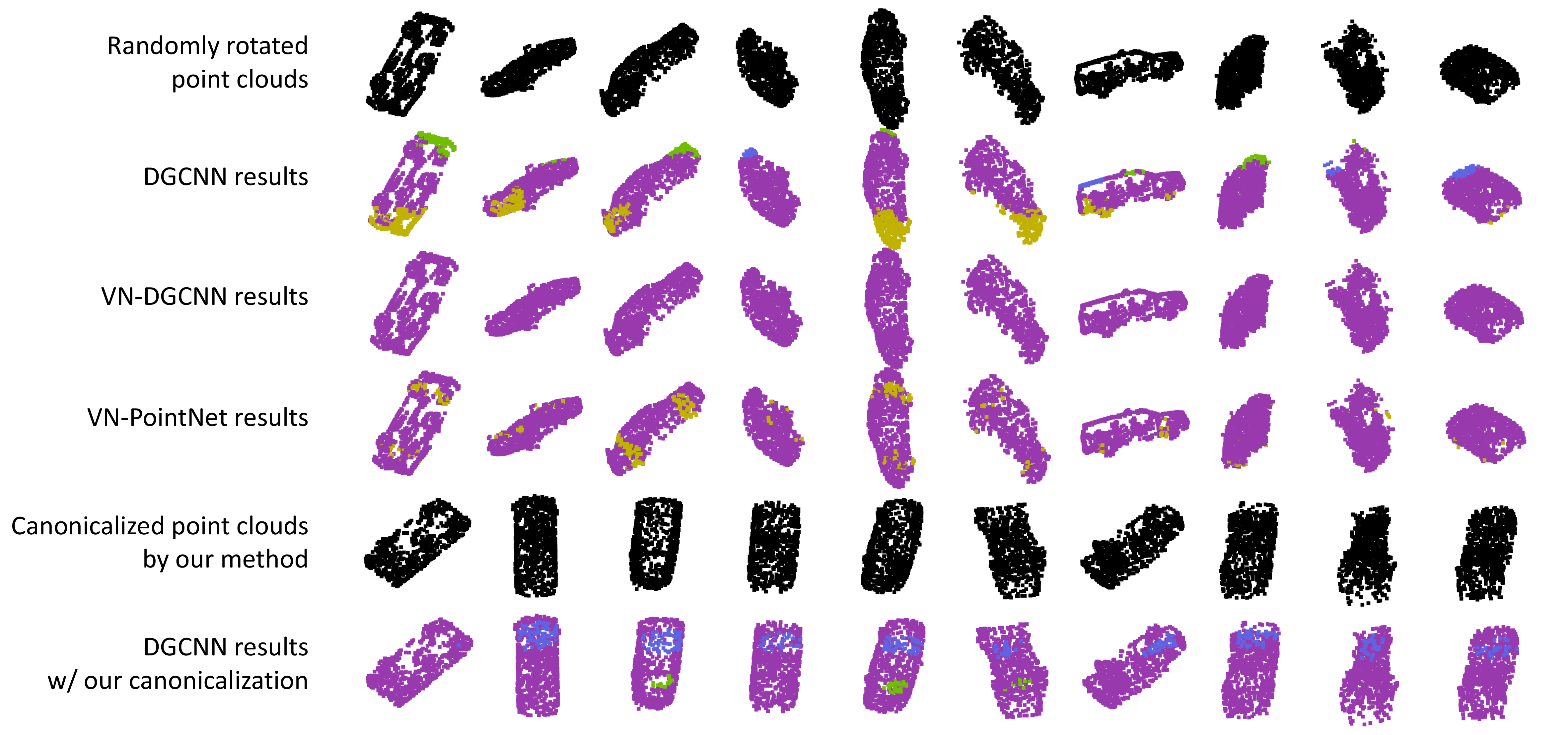}
    }
    \caption{\textbf{Qualitative visualization of part segmentations of our method (car, I/SO(3)) on the ShapeNet dataset.}
    }
    \label{fig:supp_car_seg}
\end{figure*}

\begin{figure*}[!t]
    \centering
    \resizebox{\textwidth}{!}{
    \includegraphics{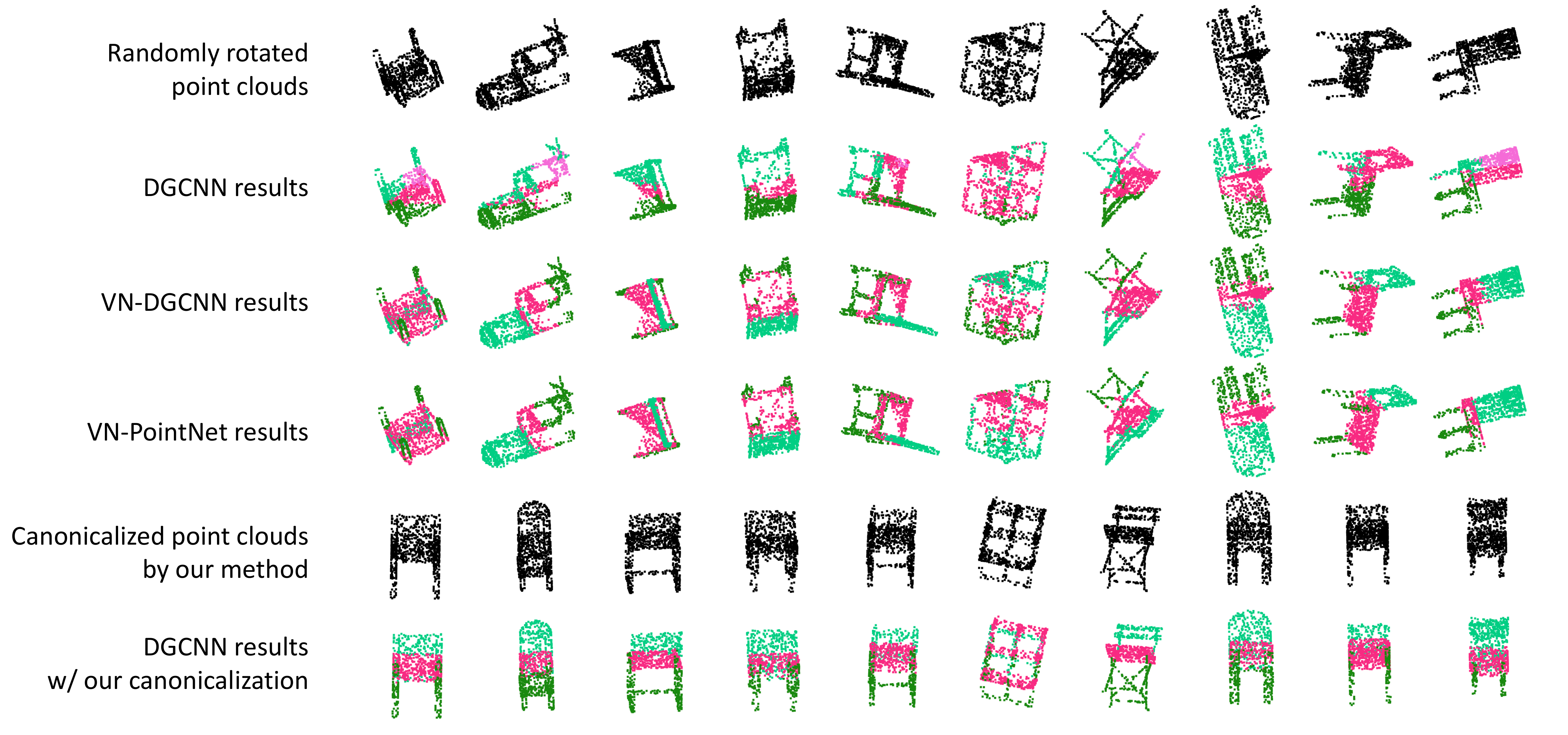}
    }
    \caption{\textbf{Qualitative visualization of part segmentations of our method (chair, I/SO(3)) on the ShapeNet dataset.}
    }
    \label{fig:supp_chair_seg}
\end{figure*}

\begin{figure*}[!t]
    \centering
    \resizebox{\textwidth}{!}{
    \includegraphics{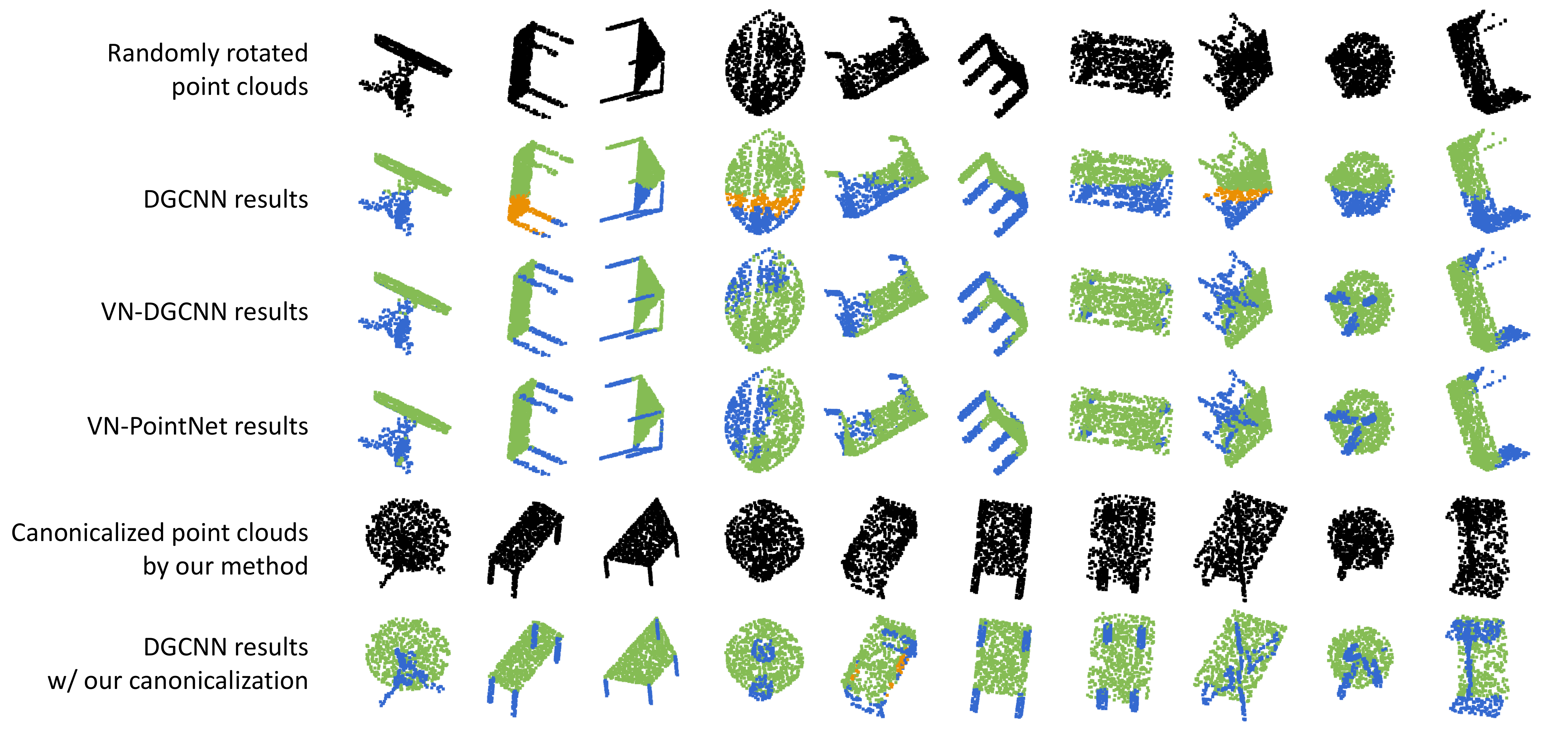}
    }
    \caption{\textbf{Qualitative visualization of part segmentations of our method (table, I/SO(3)) on the ShapeNet dataset.}
    }
    \label{fig:supp_table_seg}
\end{figure*}

In this appendix, we provide more qualitative results of our method on both characteristic orientation estimation (Figures~\ref{fig:supp_plane},\ref{fig:supp_car},\ref{fig:supp_chair},\ref{fig:supp_table}, and \ref{fig:supp_multi}) and part segmentation tasks (Figures~\ref{fig:supp_car_seg}, \ref{fig:supp_chair_seg}, and \ref{fig:supp_table_seg}).
\clearpage
\clearpage
\clearpage
\clearpage
\clearpage
\clearpage
\clearpage

\end{document}